\documentclass[conference]{IEEEtran}

\usepackage{cite}
\usepackage{amsmath,amssymb,amsfonts}
\usepackage{algorithmic}
\usepackage{graphicx}
\usepackage{textcomp}
\usepackage{xcolor}
\def\BibTeX{{\rm B\kern-.05em{\sc i\kern-.025em b}\kern-.08em
    T\kern-.1667em\lower.7ex\hbox{E}\kern-.125emX}}

\usepackage{hyperref}
\usepackage{bbm}
\usepackage{booktabs}
\usepackage{multirow}

\makeatletter
\def\ps@IEEEtitlepagestyle{%
  \def\@oddfoot{\mycopyrightnotice}%
}
\def\mycopyrightnotice{%
  \begin{minipage}{\textwidth}
  \centering \scriptsize
  Copyright~\copyright~2025 IEEE.  Personal use of this material is permitted. Permission from IEEE must be obtained for all other uses, in any current or future media, including reprinting/republishing this material for advertising or promotional purposes, creating new collective works, for resale or redistribution to servers or lists, or reuse of any copyrighted component of this work in other works.
  \end{minipage}
}
\makeatother

\begin{document}

\title{What Does an Audio Deepfake Detector Focus on? A Study in the Time Domain}
\author{
    \IEEEauthorblockN{Petr Grinberg\IEEEauthorrefmark{1}\IEEEauthorrefmark{2},
    Ankur Kumar\IEEEauthorrefmark{2}, 
    Surya Koppisetti\IEEEauthorrefmark{2},
    Gaurav Bharaj\IEEEauthorrefmark{2}}
    \IEEEauthorblockA{\textit{\IEEEauthorrefmark{1}EPFL}, Switzerland, \textit{\IEEEauthorrefmark{2}Reality Defender Inc.}, USA}
    \IEEEauthorblockA{petr.grinberg@epfl.ch, \{ankur, surya, gaurav\}@realitydefender.ai}
}

\maketitle

\begin{abstract}
Adding explanations to audio deepfake detection (ADD) models will boost their real-world application by providing insight on the decision making process. In this paper, we propose a relevancy-based explainable AI (XAI) method to analyze the predictions of transformer-based ADD models. We compare against standard Grad-CAM and SHAP-based methods, using quantitative faithfulness metrics as well as a partial spoof test, to comprehensively analyze the relative importance of different temporal regions in an audio. We consider large datasets, unlike previous works where only limited utterances are studied, and find that the XAI methods differ in their explanations. The proposed relevancy-based XAI method performs the best overall on a variety of metrics. Further investigation on the relative importance of speech/non-speech, phonetic content, and voice onsets/offsets suggest that the XAI results obtained from analyzing limited utterances don't necessarily hold when evaluated on large datasets.
\end{abstract}


\section{Introduction}
\label{sec:intro}

Recent audio deepfake detection (ADD) models are based on pre-trained self-supervised models (SSL), e.g. Wav2Vec2~\cite{baevski2020wav2vec}, which serve as black-box feature extractors but achieve strong performance on the downstream task \cite{li2024audio, tak2022automatic, lu2024one, wang2024can}. While several papers have pursued explainable AI (XAI, \cite{arrieta2020explainable}) for ADD \cite{ge2022explaining, ge2022explainable, chettri2018analysing, salvi2023towards, lim2022detecting, kwak2023voice, halpern1232020residual, zhang2023impact}, limited work exists on the interpretation of SSL models at raw waveform level in the time domain \cite{liu2024neural, li2024interpretable}. 
XAI for ADD in the time domain is necessary as the ground truth explanations are perceptual and require listening to the audios to identify the temporal regions with the artifacts. For example, Ge et. al.~\cite{ge2022explainable} studied 100 random samples for different attack types in the ASVspoof 2019 (ASV19) dataset~\cite{wang2020asvspoof} by viewing heatmaps and listening to the utterances  to conclude that artifacts may lie in the vowel region. However, such analysis may be limited to the manually inspected samples and XAI method used since XAI methods can differ in their explanations~\cite{krishna2024the}.

Despite the state-of-the-art ADD models being predominantly based on the transformer architecture \cite{vaswani2017attention}, little work has been done on transformer XAI for ADD models. Methods such as AttHear~\cite{akman2024atthear} provide some interpretation of audio transformer models, but they do not perform feature attribution, i.e., assign importance scores to the input signal timesteps. A standard approach for feature attribution in transformer models is to study the attention maps. The technique proposed in \cite{chefer2021generic} for Vision Transformer (ViT) \cite{dosovitskiy2020image} is of particular interest.
It iteratively calculates a relevancy map $R \in \mathbb{R}^{s\times s}$, where $s$ is the number of tokens in the input sequence. Given a self-attention map $A_i$ after softmax operation for $i$-th layer in the transformer, the relevancy $R$, initialized with identity matrix, is updated as:
\begin{equation}\label{eq:transformer_rule}
\begin{split}
    R_{\text{upd}} &= R_{\text{old}} + \bar{A}_i \cdot R_{\text{old}}, \text{ and }     \bar{A}_i = \mathbb{E}_h \{\left(\nabla A_i \odot A_i\right)^{+}\}.
\end{split}
\end{equation}
In \eqref{eq:transformer_rule}, $\odot$ is the Hadamard product, $\nabla A_i$ is the gradient of $A_i$ with respect to the model output score corresponding to the class of interest, and $\mathbb{E}_h$ is an average over the heads in case of multi-head self-attention. The row corresponding to \texttt{[CLS]} token is taken as the measure of relevancy for classifier decisions. {It is not straightforward to apply this method to  transformer-based ADD models such as Wav2Vec2-AASIST~\cite{tak2022automatic}, where there is no \texttt{[CLS]} token. }  
To build a transformer XAI method that directly shows the impact of each timestep in a waveform on the model outcome, we need an intelligent aggregation of relevance of the output tokens. In this regard, we pursue gradient weighted averaging and refer to the resulting method as the Gradient Average Transformer Relevancy (GATR) method.   
Our main contributions are:

\begin{enumerate}
    \item  To analyze transformer-based ADD models that take raw audio waveform inputs, we modify the relevancy maps and propose the Gradient Average Transformer Relevancy (GATR) method.
    \item We compare GATR with existing XAI methods such as Grad-CAM~\cite{kwak2023voice, halpern1232020residual, zhang2023impact, liu2024neural} and SHAP~\cite{ge2022explaining, ge2022explainable}, to study their effectiveness on a state-of-the-art SSL-based ADD classifier, Wav2Vec2-AASIST~\cite{tak2022automatic}. The methods are found to differ in the regions they highlight in an audio, so we present quantitative metrics that help to choose one method over others.
    
    \item We go beyond traditional XAI for ADD classifiers and perform hypothesis testing at a dataset-level to understand the relative importance of audio regions such as speech/non-speech, voice onsets/offsets, vowels, and consonants towards model decisions. We consider both fully spoofed and partially-spoofed utterances.
\end{enumerate}
\section{Methodology}\label{sec:method}

We consider the popular Wav2Vec2-AASIST~\cite{tak2022automatic} as an example transformer ADD model. We study the effectiveness of four XAI methods, namely Grad-CAM, DeepSHAP, GradientSHAP, and the proposed GATR method. All the considered XAI methods produce class-specific heatmaps. We use the ground truth class to get the heatmaps. We suggest a procedure to compare and analyze the output heatmaps at scale. 

\subsection{Standard XAI methods}\label{sec:method_xai_lit}
Grad-CAM~\cite{selvaraju2017grad} technique is based on gradient back-propagation from model output to the desired convolutional feature map. Wav2Vec2~\cite{baevski2020wav2vec} applies several 1-d convolution layers on the input waveform. We take the last layer here for Grad-CAM experiments. DeepSHAP and GradientSHAP~\cite{lundberg2017unified} are variants of SHAP, which is an additive feature attribution method that assigns importance based on the Shapley values~\cite{shapley1953value}.
For DeepSHAP, we follow \cite{ge2022explaining} but set the reference value with $20$ random utterances from the bona fide class in the training set instead of using both bona fide and spoof classes.
For GradientSHAP, our recipe is identical to \cite{ge2022explainable}: baselines are set to zero-vectors and there are $20$ randomly generated examples per utterance.
Similar to \cite{ge2022explaining, ge2022explainable}, we replace negative scores in SHAP with zeros. Hence, values in heatmaps from all XAI methods considered in our work are non-negative.

\subsection{Gradient Average Transformer Relevancy (GATR) Method}\label{sec:method_transformer}

In the case of ViT, the \texttt{[CLS]} token is used to choose the row from $R$, calculated via \eqref{eq:transformer_rule}, as a measure of relevancy for the classifier decisions. The Wav2Vec2-AASIST~\cite{tak2022automatic} and other modern ADD systems~\cite{lu2024one, wang2024can, martin2022vicomtech}, however, are not trained with the \texttt{[CLS]} token.
We resolve the issue by averaging the rows of $R$ across all tokens. 
We use gradient weighted average, where each row $t$ is assigned a weight $W_t$, $\forall t= 1, \dots s,$ as
\begin{equation}\label{eq:transformer_grad_average_weight}
W_t = \|(\hat{A}_{W})_t\|_2, \text{where, } \hat{A}_{W} = \mathbb{E}_h (\nabla A_N).
\end{equation}
In \eqref{eq:transformer_grad_average_weight},  
$A_N$ is the attention map for the final transformer layer and $(\hat{A}_{W})_t$ is the $t$-th row of matrix $\hat{A}_W$. The final relevancy vector $r$ is calculated as $ r = (\sum_{t=1}^{s}W_t \cdot R_t) / (\sum_{t=1}^{s} W_t)$. 
Finally, we interpolate the above relevancy vector to match the length of the input waveform $T$ since Wav2Vec2 compresses the input waveform before feeding it to the transformer.
The resulting method is referred to as Gradient Average Transformer Relevancy (GATR).
We noticed that the scores on the diagonal of $R$ are much higher than off it due to the initialization with an identical matrix. 
Thus, before taking the average, we subtract the identity matrix $I\in \mathbb{R}^{s\times s}$ from $R$.

\subsection{Comparison of different XAI methods}\label{sec:method_comparison}

\noindent \textit{ a) Faithfulness metrics:} 
Common metrics to evaluate the faithfulness of heatmaps include Average Increase (AI $\uparrow$, \cite{chattopadhay2018grad}), Average Drop (AD $\downarrow$, \cite{chattopadhay2018grad}), Average Gain (AG $\uparrow$, \cite{zhang2023opti}), and Input Fidelity (Fid-In $\uparrow$, \cite{paissan2023posthoc}). 
We use the (peak-normalized) heatmap from above XAI methods and multiply it with the input waveform to get a modified waveform. Then, Fid-In measures the change in predicted class between original and modified waveforms. We use the threshold corresponding to the equal error rate (EER) for model prediction. Similarly, AD measures the decrease in model confidence between original and modified waveforms, whereas AI and AG measure the increase in confidence.

\noindent \textit{ b) Perturbation test:} 
We conduct both positive and negative perturbation tests~\cite{chefer2021generic}. In the positive test, we noise-mask (replace with noise) timesteps that correspond to the $n\%$ highest scores in the heatmap. We then calculate the EER of the classifier on these masked utterances for $n\in [10\%, 90\%]$. 
In the negative test, the $n\%$ timesteps with the lowest scores are noise-masked. We mask with Gaussian noise, having zero mean and same variance as the input waveform, instead of zero-mask to make the perturbed sample less out-of-domain w.r.t. the original, knowing that Wav2Vec2-AASIST was trained with noise augmentations~\cite{tak2022automatic}. 
For a good explanation, high (low) scores are expected to have high (low) impact on the classifier decision and, therefore, the EER should be high (low) after the positive (negative) perturbation. We report the area under curve (AUC) for both tests. 

\noindent \textit{ c) Partial spoof test:}
Previous faithfulness metrics and perturbation test work with ground truth binary class label and do not evalaute the XAI methods in localizing the artifacts in the input audio. We propose to use partially spoofed audio that have both bona fide and spoof regions with corresponding labels ~\cite{zhang2022partialspoof}, which can serve as coarse ground-truth explanations. 
If an XAI method tracks the ground truth explanations well, it should give more importance to the spoof regions in the input audio when the classifier deems it as spoof. To evaluate this hypothesis, as well as a few others listed in Section \ref{sec:results}, we use the RCQ metric, detailed next.   

\subsection{Automated evaluation of hypotheses}\label{sec:method_rcq}

We use the Relative Contribution Quantification
(RCQ) \cite{liu2024neural} as a metric to conduct hypothesis testing at scale. RCQ  was originally proposed in \cite{liu2024neural} to understand whether a classifier focused on speech or non-speech regions. Here, we extend the metric to any set of categories $C$ relevant for evaluating a hypothesis. 
Given an utterance, we consider category $c_i\in C$ for each timestep $i\in\{0, ..., T\}$. Upon obtaining the heatmap scores $s_i$ for each timestep $i$ from a given XAI method, we can calculate the scores $S_c$ for each category $c \in C$ over a dataset of $N$ utterances as 
\begin{equation}\label{eq:rcq_score}
    S_c = \left\{{1}/{\textstyle\sum_{n=1}^{n=N}\textstyle\sum_{i=1}^{i=T}\mathbbm{1}_{c_i = c}}\right\}\textstyle\sum_{n=1}^{n=N}\textstyle\sum_{i=1}^{i=T}s_i \cdot \mathbbm{1}_{c_i = c}.
\end{equation}
The RCQ for category c is calculated as
\begin{equation}\label{eq:rcq}
    \text{RCQ}_c = 100 \cdot (S_c - S_{\text{All}})/{S_{\text{All}}}, 
\end{equation}
where $S_{\text{All}}$ is the average of scores over  all considered utterances and all timesteps, ignoring the category. If we want to investigate the model behaviour on spoof (bona fide) audio, we use spoof (bona fide) subset of the dataset for $S_{\text{All}}$ calculation.
Before calculating RCQ, we normalize all heatmaps to $[0, 1]$ to ensure that different utterances have equal contribution to the final output.
The higher the RCQ, the more important is the corresponding category, according to the XAI method.

For the partial spoof tests, we additionally calculate the Relevance Rank Accuracy (RRA) and Relevance Mass Accuracy (RMA) metrics from \cite{arras2022clevr}. The RRA measures how much of the high relevance scores lie within the ground truth region, whereas the RMA measures the ratio of high relevance scores assigned to the ground truth when compared to the entire utterance. We set the spoof regions as the ground truth.
\section{Experiments and Results}\label{sec:results}

We work with the pre-trained Wav2Vec2-AASIST model from the official open-source implementation~\cite{tak2022automatic}. We conduct all the experiments on the logical access evaluation set of ASV19 dataset~\cite{wang2020asvspoof} and In-The-Wild (ITW) dataset~\cite{muller2022does}, on which the model achieves $0.2\%$ and $10.8\%$ equal error rate (EER), respectively. For the model, ASV19 is in-domain with relatively clean audio, whereas ITW is out-of-domain with more realistic acoustic conditions and newer spoofing methods. For the partial spoof test, we use PartialSpoof dataset~\cite{zhang2022partialspoof} which is based on ASV19 but has both bona fide and spoof regions within an utterance. The model achieves $6.6\%$ EER on this dataset. Since the model is trained on peak-normalized inputs, we apply the same normalization in all experiments.

\begin{figure}[!tp]
    \centering
    \begin{minipage}{0.48\textwidth}
     \centering
     \includegraphics[width=\linewidth]{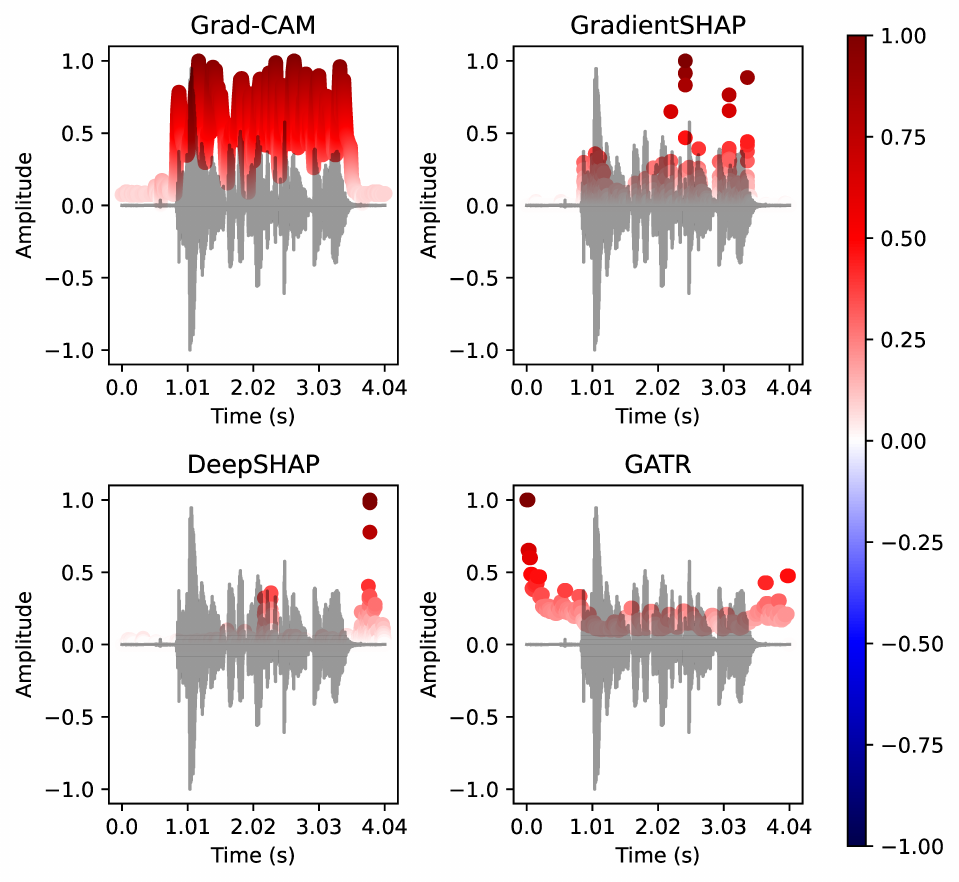}
   \end{minipage}\hfill
   \begin{minipage}{0.48\textwidth}
     \centering
     \includegraphics[width=\linewidth]{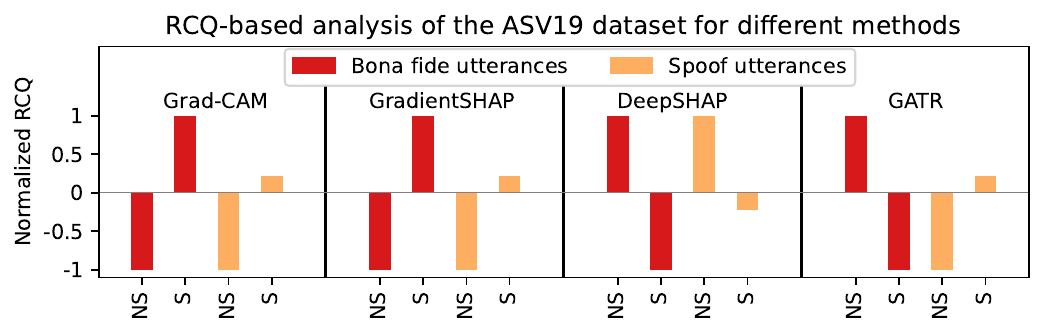}
   \end{minipage}
    \caption{(Top) Explanations from different XAI methods for bona fide \texttt{LA\_E\_4965589} utterance from the ASV19 Dataset. The color intensity and circle height indicate the heatmap score. (Bottom) Verification if the model focuses on speech (S) or non-speech (NS) regions according to the RCQs calculated for different XAI methods. 
    }
    \label{fig:utterance_example_and_rcq}
\end{figure}

\textit{Disagreement of XAI methods:}
Existing works such as \cite{liu2024neural, zhang2023impact, ge2022explainable} have shown that the ADD classifier may focus on either speech or non-speech regions.
We can use XAI to verify this. Fig. \ref{fig:utterance_example_and_rcq} (top) shows explanations from the four XAI methods for a sample utterance. We notice opposite interpretations: DeepSHAP shows that the classifier focuses mainly on non-speech region whereas GradientSHAP highlights the speech region. This is in line with observations made on {\em post-hoc} XAI methods for some other classification tasks~\cite{krishna2024the}. 

Unlike previous works, where only a handful of utterances were studied, here we report that the difference holds at scale too, when tested on the dataset-level. We obtain the speech (S) and non-speech (NS) categories for each timestep in the utterances using WebRTC voice activity detector (VAD)~\cite{url:py_webrtc, url:google_webrtc} and calculate the RCQ values. We normalize the RCQs for each XAI method separately by dividing with the maximum RCQ magnitude across all the hypothesis categories. This does not affect the relative importance of categories but allows to plot RCQs for all the XAI methods together. Fig. \ref{fig:utterance_example_and_rcq} (bottom) shows that the four XAI methods offer differing interpretations on the ASV19 dataset. For example, Grad-CAM believes speech regions are more important for the decision-making on both bona fide and spoof utterances, whereas DeepSHAP assigns more importance to the non-speech regions.

\textit{Comparison of the XAI methods:}
To better understand the real behaviour of the classifier, we need to evaluate the faithfulness of heatmaps from the different XAI methods and choose the one that offers the most consistent explanations. 
We calculate the metrics from Section \ref{sec:method_comparison} and present them in Tables \ref{tab:comparison_metrics} and \ref{tab:rcq_partial} for  the ASV19/ITW and PartialSpoof datasets, respectively. From Table \ref{tab:comparison_metrics}, the proposed GATR method is seen to perform better than the other XAI methods on most metrics.
GradientSHAP achieves superior performance in terms of AUC in the positive perturbation test and AG, but it does not follow the coarse ground truth explanations for the partially spoofed audio (Table II). In fact, from Table II, we note that all methods except GradientSHAP are able to assign high importance to the spoof regions  when the classifier deems an utterance as spoof. Through a majority-vote across the different metrics, we can choose the GATR method to analyze the model behaviour.

\begin{table}[!tp]
    \centering
    \caption{Comparison metrics from Section \ref{sec:method_comparison} on the ASV19 and ITW Datasets for all considered XAI methods. Positive perturbation test is marked as Pos. Negative -- as Neg.}
    \label{tab:comparison_metrics}
    \resizebox{\linewidth}{!}{
    \begin{tabular}{l|cc|cccc}
    \toprule
    \multirow{2}{*}{Method} &
    \multicolumn{2}{c|}{AUC-EER} & \multirow{2}{*}{AI $\uparrow$} & \multirow{2}{*}{AD $\downarrow$} & \multirow{2}{*}{AG $\uparrow$} & \multirow{2}{*}{Fid-In $\uparrow$}\\
     & Pos. $\uparrow$ & Neg.$\downarrow$ & & & & \\
    \midrule
    \textbf{ASV19} & & & & & & \\
    \midrule
    Grad-CAM  & 9.89 & 7.14 & 67.54 & 5.68 & 34.18 & 0.945\\
    GradientSHAP & \textbf{24.22} & 6.72 & 63.21 & 9.43 & \textbf{35.15} & 0.959\\
    DeepSHAP & 15.44 & 10.64 & 58.89 & 10.57 & 32.79 & 0.882\\
    GATR (ours) & 14.56 & \textbf{2.22} & \textbf{75.29} & \textbf{0.35} & 32.09 & \textbf{0.998}\\
    \midrule
    \textbf{ITW} & & & & & & \\
    \midrule
    Grad-CAM  & 32.78 & 26.48 & 60.33 & 29.96 & 44.70 & 0.736\\
    GradientSHAP & \textbf{42.41} & 18.16 & 60.67 & 26.16 & \textbf{47.10} & 0.669\\
    DeepSHAP  & 39.80 & 23.28 & 58.96 & 26.35 & 47.03 & 0.541\\
    GATR (ours) & 25.75 & \textbf{12.74} & \textbf{65.44} & \textbf{21.98} & 43.54 & \textbf{0.938}\\
    \bottomrule
    \end{tabular}
    }
\end{table}

\begin{table}[!tp]
    \centering
    \caption{Normalized RCQs for different categories, RMA, and RRA for different XAI methods on the PartialSpoof dataset. bona fide regions category is indicated as BR, spoof regions -- as SR.}
    \label{tab:rcq_partial}
    \begin{tabular}{l|cc|cc}
    \toprule
    Method & BR & SR & RMA $\uparrow$ & RRA $\uparrow$\\
    \midrule
    Grad-CAM & -0.701 & 1.000 & 0.42 & 0.43\\
    GradientSHAP & 0.701 & -1.000 & 0.40 & 0.40\\
    DeepSHAP & -0.702 & 1.000 & 0.45 & 0.44\\
    GATR (ours) & -0.703 & 1.000 & 0.44 & 0.51\\
    \bottomrule
    \end{tabular}
\end{table}

\begin{figure*}[ht]
    \centering
    \includegraphics[width=\textwidth]{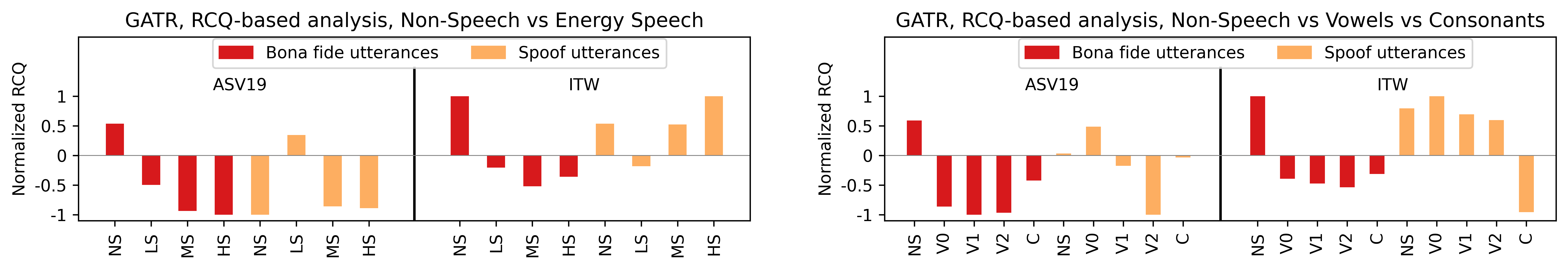}
    \caption{Normalized RCQ scores for different categories calculated on GATR heatmaps. The datasets are divided by the vertical lines. Larger normalized RCQ values indicate higher importantance of the corresponding category according to GATR. 
    Category names: S: speech, NS: non-speech, V0: unstressed vowel, V1: primary stressed vowel, V2: secondary stressed vowel, C: consonants, LS: low-energy speech, MS: middle-energy speech and HS: high-energy speech.}
    \label{fig:gatr_rcq_bars}
\end{figure*}

\textit{Automatic evaluation of hypotheses:}
We discuss two common hypotheses from the ADD literature in detail, namely the relative importance of speech compared to non-speech regions and importance of vowels compared to consonants in the speech regions. We briefly discussed the importance of silence in previous sections, which was also explored in \cite{muller2021speech, zhang2023impact}. The role of vowels was investigated in \cite{ge2022explainable} by manually analyzing a limited number of utterances, which may not hold for a larger dataset. Therefore, we evaluate these hypotheses for an entire dataset by applying RCQ on the GATR explanations. 

\textit{Speech vs Non-Speech:}
We use the WebRTC VAD to obtain speech and non-speech regions.
We further divide the speech region into low-energy (LS), middle-energy (MS), and high-energy (HS) categories to also study the importance of voice onset and offset regions, following \cite{ge2022explainable}. 
The boundaries are obtained by linearly dividing amplitude range $[0, 1]$ in the log-scale into three intervals (low, middle, and high), as was done in FastSpeech2~\cite{ren2021fastspeech}. 
The normalized RCQs for these categories are presented in the left part of the Fig. \ref{fig:gatr_rcq_bars}. A score of -1 (or 1) means that the GATR explanation assigns the lowest (or highest) importance. We find similar trend for bona fide utterances in both ASV19 and ITW datasets. Non-speech (NS) regions are the most important among all regions and voice onset region (LS) has higher importance among the low, middle and high energy speech regions. However, for spoofed utterances, we find a conflicting trend for the two datasets. Voice onset (LS) seems to be the most important with non-speech (NS) being the least for ASV19. However, LS has the lowest importance for ITW dataset with high energy (HS) regions being the most important.

\textit{Vowels vs Consonants:} 
In \cite{ge2022explainable}, the authors found that the artifacts may lie in dominant vowels regions. To verify this hypothesis at scale, we compare the importance of consonants (C), unstressed vowels (V0), primary-stressed vowels (V1), secondary-stressed vowels (V2), and non-speech (NS).
To compute RCQ scores, we obtain the text transcription for each utterance using NVIDIA Canary-1b~\cite{url:nvidia} automatic speech recognition system and then use Montreal Forced Aligner~\cite{mcauliffe2017montreal} to get phoneme-level alignment of the transcription and speech.
From Fig.~\ref{fig:gatr_rcq_bars} (right), we see that for the spoof class, the unstressed vowel (V0) is the most important, while consonant regions are the second most important speech regions on the ASV19 dataset. However, for the ITW dataset, all the vowel categories are about equally important with consonant being the least important. We skip the bonafide utterances since non-speech is the most important region for both datasets.

Thus, even though the existing literature shines some light on the decision making process of the ADD systems, the conclusions do not necessarily hold at scale for different datasets. This indicates that utterance level analysis is useful for proposing concepts but cannot be generalized to a whole dataset without additional evaluation.

\section{Conclusion}\label{sec:conclusion}

In this paper, we propose the Gradient Average Transformer Relevancy method (GATR), which facilitates interpretation for transformer-based ADD models. We compare GATR with Grad-CAM, DeepSHAP, and GradientSHAP methods, and find that they can differ significantly in their interpretations. To choose an XAI method, we conduct a faithfulness evaluation of the heatmaps at scale on two standard test sets, and report that the GATR method performs better than the others on most quantitative metrics. We then build on the GATR explanations and use the relative contribution quantification metric to test multiple hypotheses at scale, unlike previous works that rely on observations from a handful of utterances.  We notice that the non-speech play a major role in the classifier decision on bona fide data,  making it susceptible to easy adversarial attacks such as noise manipulation or the use of a denoiser. From a phonetic perspective, unstressed vowel regions are seen to have a bigger impact on the classifier when compared to other content on spoof utterances. We also highlight that some hypotheses, like focus on voice onsets/offsets, may not generalize to different datasets.

For a deeper understanding on the classifier outcomes, one direction for future research is to build ADD classifiers that are interpretable by design. Early work on this front~\cite{li2024interpretable} suggests great scope for further exploration. Even though interpretation guidance may introduce restrictions on the architecture, and may sometimes lead to worse performance, the outcomes from such models will inherently hold explanation capability.

\clearpage

\bibliographystyle{./IEEEtran}
\bibliography{./IEEEabrv, ./references}

\end{document}